\documentclass[11pt]{article}
\usepackage{coling2016}
\usepackage{times}
\usepackage{url}
\usepackage{latexsym}
\usepackage{graphicx}
\usepackage{subfigure}
\usepackage{array}
\usepackage{amsmath}
\usepackage{amssymb}
\usepackage{mathrsfs}

\title{Connecting Phrase based Statistical Machine Translation Adaptation}

\author{ Rui Wang$^{1}$, Hai Zhao$^{1}$, Bao-Liang Lu$^{1}$, Masao Utiyama$^2$ and Eiichiro Sumita$^2$\\
	$^1$ Shanghai Jiao Tong University, Shanghai, China\\
	$^2$National Institute of Information and Communications Technology, Kyoto, Japan\\
	wangrui.nlp@gmail.com, \{zhaohai, blu\}@cs.sjtu.edu.cn, \\
	\{mutiyama, eiichiro.sumita\}@nict.go.jp \\
}

\date{}

\begin{document}
\maketitle
\begin{abstract}
  Although more additional corpora are now available for Statistical Machine Translation (SMT), only the ones which belong to the same or similar domains with the original corpus can indeed enhance SMT performance directly. Most of the existing adaptation methods focus on sentence selection.  In comparison, phrase is a smaller and more fine grained unit for data selection, therefore we propose a straightforward and efficient connecting phrase based adaptation method, which is applied to both bilingual phrase pair and monolingual $n$-gram adaptation. The proposed method is evaluated on IWSLT/NIST data sets, and the results show that phrase based SMT performance are significantly improved (up to +1.6 in comparison with phrase based SMT  baseline system and +0.9 in comparison with existing methods).
\end{abstract}

\section{Introduction}
Large corpus is important for Statistical Machine Translation (SMT) training. However only the relevant additional corpora, which are also called in-domain or related-domain corpora, can enhance the performance of SMT effectively. Otherwise the irrelevant additional corpora, which are also called out-of-domain corpora, may not benefit SMT \cite{koehn-schroeder:2007:WMT}.

SMT adaptation means selecting useful part from mix-domain (mixture of in-domain and out-of-domain) data, for SMT performance enhancement. The core task in adaptation is about how to select the useful data. Existing works have considered selection strategies with various granularities, though most of them only focus on sentence-level selection \cite{axelrod-he-gao:2011:EMNLP,banerjee-EtAl:2012:PAPERS,duh-EtAl:2013:Short,hoang-simaan:2014:EMNLP2014,hoang-simaan:2014:Coling}. There is a potential problem for sentence level adaptation: different parts of a sentence may belong to different domains. That is, it is possible that a sentence is overall out-of-domain, although part of it can be in-domain. Therefore a few  works consider more granular level for selection. They build lexicon, Translation Models (TMs), reordering models or Language Models (LMs) to select fragment or directly adapt the models \cite{Bellegarda200493,deng-xu-gao:2008:ACLMain,moore-lewis:2010:Short,foster-goutte-kuhn:2010:EMNLP,mansour-ney:2013:NAACL-HLT,carpuat-EtAl:2013:ACL2013,chen-foster-kuhn:2013:NAACL-HLT,chen-kuhn-foster:2013:ACL2013,sennrich-schwenk-aransa:2013:ACL2013,mathur-venkatapathy-cancedda:2014:Coling,Shi2015136}.  One typical example of these methods is to train two Neural Network (NN) models (one from in-domain and the other from out-of-domain) and penalize the sentences/phrases similar to out-of-domain corpora \cite{duh-EtAl:2013:Short,joty-EtAl:2015:EMNLP2,durrani2015using}.  As we know, Phrase based SMT (PBSMT) mainly contains two models: translation model and LM, whose components are bilingual phrase pairs and monolingual $n$-grams. Meanwhile, most of the above methods enhance SMT performance by adapting single specific model.

Instead of focusing on sentence selection or single model adaptation, we propose a phrase adaptation method, which is applied to both bilingual phrase pair and monolingual $n$-gram selection. It is based on a linguistic observation that the translation hypotheses of a phrase-based SMT system are concatenations of phrases from Phrase Table (PT), which has been applied to LM growing \cite{wang-EtAl:2014:EMNLP20142}. As a straightforward linear method, it is much efficient in comparison with NN based non-linear methods.

The remainder of this paper is organized as follows. Section \ref{sec:con} will introduce the connecting phrase based adaptation method. The size of adapted connecting phrase will be tuned in Section \ref{sec:adp-size}. Empirical results will be shown in Section \ref{sec:exp}. We will discuss the methods and conduct some extension experiments in Section \ref{sec:discussion}. The last section will conclude this paper.

\section{Connecting Phrase based Adaptation}
\label{sec:con}
Suppose that two phrases `\emph{would like to learn}' and `\emph{Chinese as second language}' are in in-domain PT. In decoding, these two phrases may be connected together as `\emph{would like to learn Chinese as second language}". The phrases `\emph{would like to learn Chinese}' or `\emph{learn Chinese as second language}' may be outside in-domain PT/LM, but they may possibly be in  out-of-domain PT/LM. Traditionally their translation probabilities are only calculated by the combination of probabilities from in-domain PT/LM. For the proposed methods, the translation probabilities of connecting phrases from out-of-domain corpus are estimated by real corpus directly. If we can add these connecting phrases with their translation probabilities, which may be useful in decoding, into in-domain bilingual (together with source part phrases) PT or monolingual LM, they may help improve SMT.

Note that connecting phrases are generated from in-domain PT, it is necessary to  check if these in-domain connecting phrases actually occur in out-of-domain PT/LM. Connecting phrases can occur in decoding by combining two phrases from in-domain PT.

Let $w_a^b$ be a phrase starting from the $a$-th word and ending with the $b$-th word, and  $\gamma w_a^b\beta$ be a phrase including $w_a^b$ as a part of it, where $\gamma$ and $\beta$ represent any word sequence or none. An $i$-gram phrase $w_1^kw_{k+1}^i$ ($1 \le k \le i-1$) is a connecting phrase\footnote{We are aware that connecting phrases can be applied to three or more phrases. Experimental results show that using more than two connecting phrases cannot further improve the performance, so only two connecting phrases are applied.}  \cite{wang-EtAl:2014:EMNLP20142}, if

1) $w_1^k$ is  right (rear) part of one phrase $\gamma w_1^k$ in in-domain PT, and

2) $w_{k+1}^i$ is left (front) part of one phrase $w_{k+1}^i\beta$ in in-domain PT.

For example, let `\emph{a b c d}' be a 4-gram phrase, it is a connecting phrase if at least one of the following conditions holds:

1) `$\gamma$ \emph{a}'  and `\emph{b c d} $\beta$' are in phrase table, or 

2) `$\gamma$ \emph{a b}' and `\emph{c d} $\beta$' are in phrase table, or 

3) `$\gamma$ \emph{a b c}' and `\emph{d} $\beta$' are in phrase table. 

For a phrase pair ($F$,  $E$) in out-of-domain PT, there are four cases: a) Both $F$ and $E$, b) either $F$ or $E$, c) only $F$, d) only $E$ are/is connecting phrase(s). We empirically evaluate the performance of these four cases and the results show that a) gains the highest BLEU, so it is adopted at last. For an $n$-gram LM, we only consider target side information.

\section{Adapted Phrase Size Tuning}
\label{sec:adp-size}
A lot of connecting phrases are generated in the above way. We propose two methods to rank these phrases and only  the top ranked ones are added into in-domain PT/LM.

\subsection{Occurring Probability based Tuning}
The potential Occurring Probability (OP) of a source phrase $P_{op}(F)$ and  $P_{op}(E)$ are defined as,

\begin{equation}
		\label{eq:e}
		\begin{aligned}
			P_{op}(F) =  
			\sum_{k=1}^{p-1}(\sum_{\beta}P_{s}(\beta f_1^k) \times \sum_{\gamma}P_{s}(f_{k+1}^p\gamma)),  \\
			P_{op}(E) =   \sum_{k=1}^{q-1}(\sum_{\beta}P_{t}(\beta e_1^k) \times \sum_{\gamma}P_{t}(e_{k+1}^q\gamma)), \nonumber
		\end{aligned}
\end{equation}%
respectively, where $P_s$ (for source phrase $f_1^p$) or $P_t$ (for target phrase $e_1^q$) is calculated using source or target monolingual LM trained from in-domain corpus.

The $P_{op}(F, E)$ of a connecting phrase pair $(F, E)$ in SMT decoding is defined as  $P_{op}(F) \times P_{op}(E)$. $P_{op}(F, E)$ is used to rank connecting phrase pairs. For target LM, only $P_{op}(E)$ is used to rank connecting $n$-gram \cite{wang-EtAl:2014:EMNLP20142}.

\subsection{NN based Tuning}
\label{sec:nn-tune}
The basic hypothesis of NN based adaptation is: two NN models (translation model as NNTM or LM as NNLM),  one from in-domain and one from out-of-domain are trained. Taking NNTM as example, for a phrase pair $(F, E)$ relevant with in-domain ones, the translation probabilities $P_{in}(E|F)$ by $NNTM_{in}$ should be larger and $P_{out}(E|F)$ by $NNTM_{out}$ should be lower. This hypothesis is partially motivated by \cite{axelrod-he-gao:2011:EMNLP}, which use bilingual cross-entropy difference to distinguish in-domain and out-of-domain data.
	
The translation probability of a phrase-pair is estimated as,

\begin{equation}
	\label{cstm}
		P(E|F) = P(e_1,...,e_q|f_1,...,f_p),
\end{equation}%
where $f_s$ ($s \in  [1,p]$) and $e_t$ ($t \in [1,q]$) are source and target words, respectively. Originally,

\begin{equation}
	\label{cstm}
	P(e_1,...,e_q|f_1,...,f_p) = \\ \\
	\prod_{k=1}^{q} P(e_k|e_{1},...,e_{k-1},f_1,...f_p).
\end{equation}%

The structure of NN based translation model is similar with Continuous Space Translation Model (CSTM) \cite{schwenk:2012:POSTERS}. For the purpose of adaptation, the dependence between target words is dropped\footnote{We have also empirically compared the performance of using NN with target word dependence and the results show not so well.} and the probabilities of different length target phrase are normalized. For an incomplete source phrase, i.e. with less than seven words, we set the projections of the missing words to zero. The normalized translation probability $Q(E|F)$ can be approximately computed by the following equation,

\begin{equation}
	\label{cstm}
		Q(E|F) \approx \sqrt[q]{\prod_{k=1}^{q} P(e_k|f_1,...f_p)}.
\end{equation}%
			
Finally, the minus $D_{minus}(E|F)$ between $Q_{in}(E|F)$ and $Q_{out}(E|F)$ are used to rank connecting phrase pairs from out-of-domain PT,

\begin{equation}
	\label{minus}
	D_{minus}(E|F) =  Q_{in}(E|F) - Q_{out}(E|F).
\end{equation}

For monolingual $n$-gram tuning, two NNLMs (in and out) are trained, and

\begin{equation}
\label{minus}
D_{minus}(E) = Q_{in}(E) - Q_{out}(E), 
\end{equation}%
where $Q_{in}(E)$ and $Q_{out}(E)$ are corresponding probabilities from in-domain and out-of-domain NNLMs,  are used for $n$-gram ranking.

Beside for connecting phrases size tuning, this NN based method can also be applied to phrase adaptation directly, which is similar as other NN based adaptation methods, such as \cite{duh-EtAl:2013:Short} for sentence selection and \cite{joty-EtAl:2015:EMNLP2} for joint model adaptation. In addition, the translation probabilities of connecting phrases calculated by NN can also be used to enhance SMT, and the experimental results will be shown in Section \ref{sec:nn}.

\subsection{Integration into SMT}
The thresholds of $P_{op}$ and $D_{minus}$ are tuned using development data. Selected phrase pairs are added into the in-domain PT. Because they are not so useful as the in-domain ones, a penalty score is added. For in-domain phrase pairs, the penalty is set as 1; for the out-of-domain ones the penalty is set as $e$ (= 2.71828...). Other phrase scores (lexical weights et. al.) are used as they are. This penalty setting is similar with \cite{bisazza2011fill}. Penalty weights will be further tuned by MERT \cite{och:2003:ACL}.  The phrase pairs in re-ordering model are selected using the same way as PT. The selected monolingual $n$-grams are added to the original LM, and the probabilities are re-normalized by SRILM \cite{Stolcke2002,citeulike:11852590}.

\section{Experiments}
\label{sec:exp}
\subsection{Data sets}
The proposed methods are evaluated on two data sets (the projects will be released online in the final paper). 1) IWSLT 2014 French (FR) to English (EN) corpus is used as in-domain data and dev2010 and test2010/2011 \cite{Jan:Niehues}, are selected as development (dev) and test data, respectively. Out-of-domain corpora contain Common Crawl, Europarl v7, News Commentary v10 and United Nation (UN) FR-EN parallel corpora\footnote{It is available at \url{http://statmt.org/wmt15/translation-task.html}}. 2) NIST 2006 Chinese (CN) to English corpus\footnote{\url{http://www.itl.nist.gov/iad/mig/tests/mt/2006/}} is used as in-domain corpus, which follows the setting of \cite{wang-EtAl:2014:P14-2} and mainly consists of news and blog texts. Chinese to English UN data set (LDC2013T06) and NTCIR-9 \cite{goto:2011} patent data are used as out-of-domain bilingual (Bil) parallel corpora. The English patent data in NTCIR-8 \cite{Fujii10overviewof} is also used as additional out-of-domain monolingual (Mono) corpus. NIST Eval 2002-2005 and NIST Eval 2006 are used as dev and test data, respectively.

			\begin{table}[htbp]
				\begin{center}
					\begin{tabular}{l|r|r }
						\hline
						\hline
						IWSLT FR-EN & Sentences & Tokens \\
						\hline
						in-domain  & 178.1K & 3.5M \\
						out-of-domain & 17.8M &  450.0M\\
						dev & 0.9K & 20.1K \\
						test2010 & 1.6K & 31.9K\\
						test2011 & 1.1K & 21.4K\\
						\hline
						\hline
						NIST CN-EN & Sentences  & Tokens\\
						\hline
						in-domain  & 430.8K & 12.6M\\
						out-of-domain (Bil) & 8.8M & 249.4M\\
						out-of-domain (Mono) & 33.7M & 1.0B\\
						dev (average of four) & 4.4K & 145.8K\\
						test (average of four) & 1.6K & 46.7K\\
						\hline
						\hline
					\end{tabular}
				\end{center}
				\caption{\label{tab:data} Statistics on data sets (`B' for billions). }
			\end{table}

\subsection{Common Setting}
The basic settings of IWSLT-2014 FR to EN and NIST-06 CN to EN phrase based translation baseline systems are followed. 5-gram interpolated KN \cite{10.1109/ICASSP.1995.479394} LMs are trained. Translation performances are measured by case-insensitive BLEU \cite{Papineni:2002:BMA:1073083.1073135}  with significance test \cite{koehn:2004} and METEOR \cite{denkowski:lavie:meteor-wmt:2014}. MERT \cite{och:2003:ACL} (BLEU based) is run three times for each system and the average BLEU/METEOR scores are recorded.  4-layer CSTM \cite{schwenk:2012:POSTERS} are applied to NN translation models: phrase length limit is set as seven, shared projection layer of dimension 320 for each word (that is 2240 for seven words), projection layer of dimension 768, hidden layer of dimension 512. The dimensions of input/output layers for both in/out-of-domain CSTMs follows the size of vocabularies of source/target words from in-domain corpora. That is, 72K/57K for IWSLT 149/112K for NIST. Since out-of-domain corpora are huge, part of them are resampled (resample coefficient 0.01 for IWSLT and NIST).

Several related existing methods are selected as baselines\footnote{We are aware that there are various SMT adaptation works such as \cite{deng-xu-gao:2008:ACLMain,joty-EtAl:2015:EMNLP2}. However, there does not exist a commonly used evaluation corpus for this task, and either detailed implementations or experimental settings are absent for most published works.}: \cite{koehn-schroeder:2007:WMT}'s method for using two (in and out-of-domain) TMs and LMs together, entropy based method for TM \cite{ling2012entropy} and LM \cite{DBLP:journals/corr/cs-CL-0006025} adaptation (pruning),  \cite{duh-EtAl:2013:Short} for NNLM based sentence adaptation, \cite{sennrich:2012:EACL2012} for TM weights combination,  \cite{bisazza2011fill} for TM fill-up and \cite{hoang-simaan:2014:EMNLP2014} for sentence and TM adaptation. In Table Tables \ref{tab:results} and \ref{tab:nist-results}, `in-domain', `out-of-domain' and `mix-domain' indicate training all models using corresponding corpora, `in+NN' indicates applying NN based adaptation directly for all phrases, and `in+connect' indicates adding all connecting phrases and $n$-grams to in-domain PT and LM, respectively. For tuning methods, `in+connect+OP/NN' indicates tuning connecting phrase pairs and $n$-grams using Occurring Probability (OP) and NN, respectively. Only the best performed results (for both the baselines and proposed methods) on development data are chosen to be evaluated on test data.

\begin{figure}[htbp]
	\centering
	\includegraphics[width=0.6\textwidth]{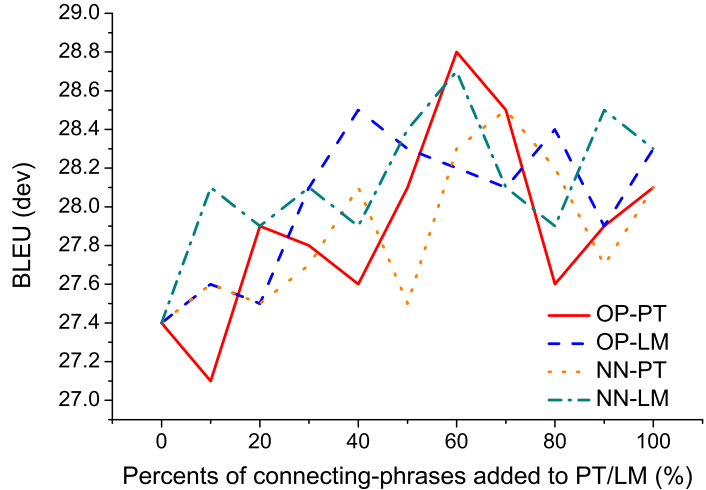}
	\caption{\label{fig:tuning} Connecting phrases size tuning on IWSLT.}
\end{figure}

\begin{table}[htbp]
	\begin{center}
		\begin{tabular}{l|rrllll }
			\hline
			\hline
			Methods & PT Size & LM Size& BLEU & METEOR &BLEU&METEOR\\
			&  & & test10 & test10 & test11& test11\\
			\hline
			in-domain  & 9.8M& 7.9M&31.94 & 34.07 &29.16&32.34\\
			out-of-domain  & 759.0M& 497.4M&27.34 & 32.22 &23.80& 30.48\\
			mix-domain & 765.4M& 503.1M&30.07&  33.19& 26.42& 31.06\\
			\hline
			Koehn's method &N/A&N/A& 32.42& 34.32& 29.41 & 32.41\\
			entropy method &247.8M&146.1M& 32.54& 34.12& 29.23 & 32.17\\
			Duh's method & 765.4M&  271.0M&\textbf{32.65}& 34.31 &29.18& 32.57\\
			Sennrich's method & 765.4M&  503.1M&32.41&  34.32 &\textbf{29.67}& \textbf{32.71}\\
			Bisazza's method & 765.4M&  503.1M&32.24& 34.28 &29.35& 32.53\\
			Hoang's method &177.3M& 212.6M&32.57&\textbf{34.45}&29.47& 32.62 \\
			\hline
			in+NN & 296.8M& 156.2M&32.54& 34.25 &29.67& 32.68\\
			in+connect &  184.5M  & 133.8M & 33.26+ & 34.60 &30.07&32.89 \\
			in+connect+OP  & 122.0M & 53.5M&\textbf{33.53}++&  \textbf{34.77} &30.25&32.91\\
			in+connect+NN  & 141.3M & 80.3M&32.91& 34.56 &\textbf{30.32}+&\textbf{33.17}\\
			\hline
			\hline					
		\end{tabular}
	\end{center}
	\caption{\label{tab:results} IWSLT FR-EN Results. \textquotedblleft $++$\textquotedblright: BLEU significantly better than corresponding the best performed baseline (in \textbf{bold}) at level $\alpha = 0.01$, \textquotedblleft $+$\textquotedblright :  $\alpha = 0.05$. Koehn's method uses two TMs and LMs, so their size is hard to tell.}
\end{table}

\begin{table}[htbp]
	\begin{center}
		\begin{tabular}{l|rrll }
			\hline
			\hline
			Methods & PT Size& LM Size & BLEU  & METEOR\\
			\hline
			in-domain  &27.2M& 23.9M&32.10&  29.29\\
			out-of-domain  & 365.8M&  1.2B&27.85& 22.48 \\
			mix-domain &370.9M& 1.2B&31.37&  28.80\\
			\hline
			Koehn's method &N/A&N/A& 31.93& 29.32\\
			entropy method &165.3M&279.5M& 32.29& 29.17\\
			Duh's method & 160.5M& 519.3M&\textbf{32.51} &  29.36\\
			Sennrich's method & 370.9M&  1.2B&32.36& 
			\textbf{29.88}\\
			Bisazza's method & 370.9M& 1.2B&32.15 &   29.72\\
			Hoang's method &153.2M&378.6M&32.50&29.73 \\
			\hline
			in+NN & 187.6M& 394.1M&32.82+& 30.23\\
			in+connect& 142.6M  &298.1M& 32.63 & 29.97\\
			in+connect+OP  &  92.6M& 208.7M&32.76& \textbf{30.63}\\
			in+connect+NN   &113.6M& 142.1M&\textbf{33.23}++& 30.54\\
			\hline
			\hline				
		\end{tabular}
	\end{center}
	\caption{\label{tab:nist-results} NIST-06 CN-EN Results.}
\end{table}

\subsection{Results and Analysis}
\label{sec:results}
			
For all ranked connecting phrase pairs and $n$-grams, we empirically add different sized (top) parts of them into PT/LM for size tuning. Figure \ref{fig:tuning} shows performances of the proposed tuning methods on IWSLT development data set. The results show that adding connecting phrases can enhance SMT performance in most of cases. Meanwhile, the tuned connecting phrases, which are parts of the whole, gain more BLEU improvement. They are considered as the most useful connecting phrases and evaluated on the test data sets.

Tables \ref{tab:results} and \ref{tab:nist-results} shows that directly using `out-of-domain' or `mix-domain' data will cause SMT performances decrease in comparison with `in-domain' data. Adding connecting phrases will enhance SMT performances and the proposed tuning method can further increase SMT performances significantly (up to +1.6 BLEU in IWSLT task and +1.1 in NIST task) and outperform the existing methods (up to +0.9 BLEU in IWSLT task and +0.7 in NIST task). The NN method performs better as a tuning method than as a direct adaptation method.

\section{Discussions}
\label{sec:discussion}

\subsection{Individual Model Analysis}
\label{sec:imp}
Most of the existing methods focus on single model adaptation, however the proposed connecting phrase method can apply to both TM and LM. So it seems a little unfair to compare the existing methods with our methods. To compare with them in a more fair way, we show the performance of individual model in Tables \ref{tab:lm-adp} and \ref{tab:tm-adp} for IWSLT tasks. Similar as the previous experiments, only the best performed system on development data of each method is evaluated on the test data.

\begin{table}[htbp]
	\begin{center}
		\begin{tabular}{l|rll }
			\hline
			\hline
			Methods  & LM Size& BLEU & BLEU \\
			&  &  test10 & test11 \\
			\hline
			in-domain  & 7.9M&31.94 & 29.16 \\
			out-of-domain  & 497.4M& 31.01& 27.42\\
			mix-domain & 503.1M& 32.23& 28.42\\
			\hline
			Koehn's method &N/A & 32.34& 29.10\\
			entropy method &146.1M& 32.31& 29.24\\
			Duh's method & 271.0M& 32.65&  29.18\\
			\hline
			in+NN & 156.2M&32.66&  29.38 \\
			in+connect& 133.8M & 32.78&   29.32\\
			in+connect+OP  & 53.5M& \textbf{32.95}&  29.45\\
			in+connect+NN  & 80.3M& 32.56& \textbf{29.78}\\
			\hline
			\hline					
		\end{tabular}
	\end{center}
	\caption{\label{tab:lm-adp} IWSLT FR-EN results on LM adaptation.}
\end{table}

\begin{table}[htbp]
	\begin{center}
		\begin{tabular}{l|rll }
			\hline
			\hline
			Methods  & PT Size& BLEU & BLEU \\
			&  &  test10 & test11 \\
			\hline
			in-domain  &9.8M &31.94 & 29.16 \\
			out-of-domain  &759.0M & 28.62& 24.56\\
			mix-domain & 765.4M& 29.56& 26.78\\
			\hline
			Koehn's method &N/A & 31.97& 29.21\\
			entropy method & 247.8M&32.43 & 28.73\\
			Sennrich's method & 765.4M& 32.41&  29.67\\
			Bisazza's method & 765.4M& 32.24&  29.35\\
			Hoang's method &177.3M&32.46&29.63 \\
			\hline
			in+NN & 296.8M& 32.31&   29.63\\
			in+connect& 184.5M & 32.87&   29.48\\
			in+connect+OP  & 122.0M& \textbf{33.05}& 29.77 \\
			in+connect+NN  & 141.3M& 32.73& \textbf{29.89}\\
			\hline
			\hline					
		\end{tabular}
	\end{center}
	\caption{\label{tab:tm-adp} IWSLT FR-EN results on TM adaptation.}
\end{table}

As shown in Tables \ref{tab:lm-adp} and \ref{tab:tm-adp}, the proposed methods outperform existing methods in individual model performance (up to +0.3 BLEU in LM task and +0.6 BLEU in TM task for test10 and +0.5 BLEU in LM task and +0.2 BLEU in TM task for test11). Another observation is that  adding out-of-domain data into TM hurt SMT system more seriously than LM (-0.9 BLEU in LM task versus -3.4 BLEU in TM task for test10 and -1.7 BLEU in LM task versus -4.6 BLEU in TM task for test11).

\subsection{Manual Example}
\label{sec:Manual}
Some adapted phrase examples of IWSLT FR-EN task are in Table \ref{tab:example}. For NN based method (direct apply NN in adaptation), some phrases with similar meaning are adapted, such as \emph{third world countries} and \emph{developing countries}. For connecting phrase method, some phrases which are combination of phrases are adapted, such as \emph{the reason} and \emph{why I like} form \emph{the reason why I like}.

\begin{table}[htbp]
	\begin{center}
		\begin{tabular}{l|l|l|l }
			\hline
			\hline
			Methods & Source Phrases &   Original Target Phrases & Adapted Phrases\\
			\hline 
			NN &   \emph{les pays en } & \emph{1. developing countries} & \emph{1. developing countries}\\
			&\emph{voie de d{\'e}veloppement} & \emph{2. the developing countries} & \emph{2. third world countries}\\
			& & \emph{3. all developing countries} & \emph{3. countries in the} \\
			& & &	\emph{ developing world}\\
			\hline
			Connect & \emph{la raison pour} & \emph{1. the reason I want} & \emph{1. the reason why I like}\\
			& \emph{laquelle je tiens} & \emph{2. why I like} & \emph{2. the reason I want}\\
			& & \emph{3. I therefore wish} & \emph{3. the reason I would like}\\
			
			\hline
		\end{tabular}
	\end{center}
	\caption{\label{tab:example} Some examples of adapted phrases, which are ranked by translation probabilities.}
\end{table}

\subsection{Efficiency Comparison}
\label{sec:efficiency}
Table \ref{tab:time} shows the adaptation time of each method\footnote{Koehn and Sennrich's method is just for model combination, so we do not compare with it.} on IWSLT task. The proposed methods show significant advantage over others, and NN based methods are very time consuming.

\begin{table}[htbp]
	\begin{center}					
		\begin{tabular}{l|r }	
			\hline
			\hline
			Methods & Adaptation Time \\
			\hline
			entropy method &  12 hours \\
			Duh's method &  7 days \\
			Bisazza's method & 6 hours\\
			Hoang's method & 34 hours\\
			in+NN & 10 days\\
			in+connect & \textbf{2 hours} \\
			in+connect+OP  &  3 hours\\
			in+connect+NN   & 3 days\\	
			\hline
			\hline
		\end{tabular}
	\end{center}
	\caption{\label{tab:time} Efficiency comparison (CPU time) on IWSLT.}
\end{table}

\subsection{Adding NN Probabilities}
\label{sec:nn}

As mentioned in Section \ref{sec:nn-tune}, NN can be used to predict the translation probabilities of bilingual phrase pairs and the occurring probabilities of monolingual $n$-grams. The minus $D_{minus}$ between in-domain NN probabilities $Q_{in}$ and out-of-domain NN probabilities $Q_{out}$  are used to judge whether a phrase (pair) is similar with the in-domain ones. Meanwhile, these in-domain NN probabilities $Q_{in}$ themselves are also useful information. In the previous sections, the adapted phrase pairs are added into original PT or LM with their own probabilities. In this subsection, the $Q_{in}$ of adapted and original phrases are also adopted in SMT decoding. That is, $Q_{in} (E|F)$ is added as a feature for adapted and original phrase pairs in PT and $Q_{in} (E)$ of adapted and original $n$-grams are interpolated with $n$-gram LM probabilities.

\begin{table}[htbp]
	\begin{center}
		\begin{tabular}{l|rrrr }
			\hline
			\hline
			Methods & PT Size & LM Size& BLEU & BLEU \\
			& & & without  $Q_{in}$ & with  $Q_{in}$ \\			\hline
			in-domain  & 9.8M& 7.9M&31.94 &  32.34\\
			\hline
			in+NN & 296.8M& 156.2M&32.54& 32.48\\
			in+connect &  184.5M  & 133.8M & 33.26 & 33.45\\
			in+connect+OP  & 122.0M & 53.5M&33.53& \textbf{33.67}\\
			in+connect+NN  & 141.3M & 80.3M&32.91& 33.12\\
			\hline
			\hline					
		\end{tabular}
	\end{center}
	\caption{\label{tab:nn-add} IWSLT FR-EN Results.}
\end{table}	

The results in Table \ref{tab:nn-add} show that the NN feature can enhance SMT performance slightly. Although this is not our main contribution, it shows the NN method cannot only be applied to phrase pair and $n$-gram adaptation, but also to probability estimation.

\section{Conclusion}
In this paper, we propose a straightforward connecting phrase based SMT adaptation method. Two model size tuning methods, NN and occurring probability are proposed to discard less reliable connecting phrases. The empirical results in IWSLT French to English and NIST Chinese to English translation tasks show that the proposed methods can significantly outperform a number of the existing SMT adaptation methods in both performance and efficiency. We also show some empirical results to discuss where does the SMT improvement come from by individual model  and manual example analysis.

\bibliography{thesis}
\bibliographystyle{acl}

\end{document}